\documentclass[letterpaper]{article} 
\usepackage{aaai25}  
\usepackage{times}  
\usepackage{helvet}  
\usepackage{courier}  
\usepackage[hyphens]{url}  
\usepackage{graphicx} 
\usepackage{natbib}  
\usepackage{caption} 
\usepackage{algorithm}
\usepackage{algorithmic}
\usepackage{amsmath,amsfonts}
\usepackage{subcaption}
\usepackage{color}
\usepackage{booktabs}

\usepackage{newfloat}
\usepackage{listings}
\DeclareCaptionStyle{ruled}{labelfont=normalfont,labelsep=colon,strut=off} 
\floatstyle{ruled}
\newfloat{listing}{tb}{lst}{}
\floatname{listing}{Listing}
\title{An Information-theoretic Multi-task Representation Learning Framework for Natural Language Understanding}

\author{Dou Hu\textsuperscript{\rm 1,\rm 2}, Lingwei Wei\textsuperscript{\rm 1}\thanks{Corresponding author.}, Wei Zhou\textsuperscript{\rm 1}, Songlin Hu\textsuperscript{{\rm 1},{\rm 2}}\footnotemark[1] 
}

\affiliations{
    \textsuperscript{\rm 1} Institute of Information Engineering, Chinese Academy of Sciences \\
    \textsuperscript{\rm 2} School of Cyber Security, University of Chinese Academy of Sciences  \\
    \{hudou, weilingwei, zhouwei, husonglin\}@iie.ac.cn \\
}

\usepackage{multirow}
\usepackage{arydshln}
\usepackage{amssymb}

\begin{document}

\maketitle 

\begin{abstract}
This paper proposes a new principled multi-task representation learning framework (InfoMTL) to extract noise-invariant sufficient representations for all tasks. It ensures sufficiency of shared representations for all tasks and mitigates the negative effect of redundant features, which can enhance language understanding of pre-trained language models (PLMs) under the multi-task paradigm. Firstly, a shared information maximization principle is proposed to learn more sufficient shared representations for all target tasks. It can avoid the insufficiency issue arising from representation compression in the multi-task paradigm. Secondly, a task-specific information minimization principle is designed to mitigate the negative effect of potential redundant features in the input for each task. It can compress task-irrelevant redundant information and preserve necessary information relevant to the target for multi-task prediction. Experiments on six classification benchmarks show that our method outperforms 12 comparative multi-task methods under the same multi-task settings, especially in data-constrained and noisy scenarios. Extensive experiments demonstrate that the learned representations are more sufficient, data-efficient, and robust. 
\end{abstract}

\begin{figure*}[t]
    \centering
    \includegraphics[width=0.99\linewidth]{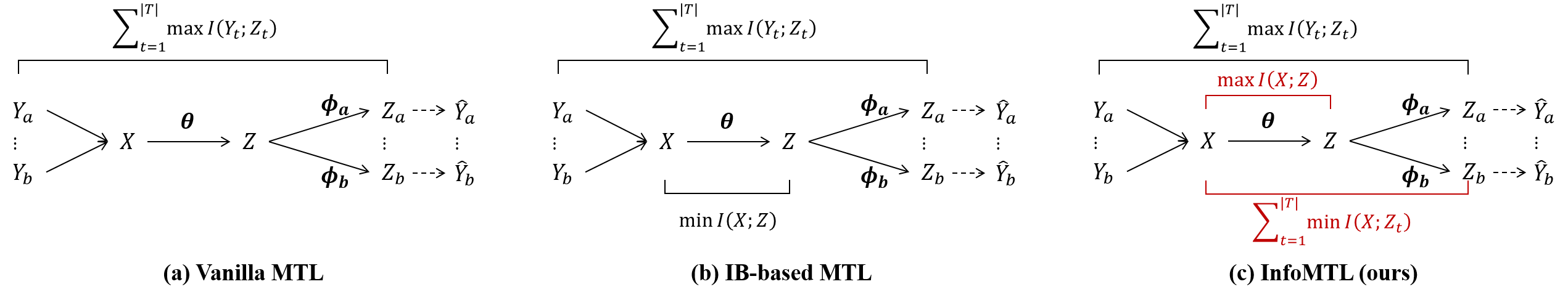}
    \caption{Comparison of different learning principles under Markov constraints in MTL paradigm. Given the input variable X, shared representations $Z$, task-specific output representations $Z_t$, and the prediction variable $\hat{Y}_t$, the Markov chain for each task $t$ is $Y_t \rightarrow X \rightarrow Z \rightarrow Z_t \rightarrow \hat{Y}_t$.}
    \label{fig:compare}
\end{figure*}

\section{Introduction}

Multi-task learning (MTL) has become a promising paradigm in deep learning to obtain language representations from large-scale data \cite{DBLP:conf/acl/LiuHCG19}.
By leveraging supervised data from related tasks, multi-task learning approaches reduce expensive computational costs and provide a shared representation which is also more efficient for learning over multiple tasks \cite{DBLP:conf/iclr/0002ZR20,royer2023scalarization}.  
 
Most works \cite{DBLP:conf/cvpr/KendallGC18,DBLP:conf/cvpr/ChennupatiSYR19,DBLP:conf/cvpr/LiuJD19,DBLP:conf/nips/YuK0LHF20,DBLP:conf/iclr/LiuLKXCYLZ21,DBLP:journals/tmlr/LinYZT22} on MTL mainly focus on balancing learning process across multiple tasks such as loss-based and gradient-based methods.  
However, in real-world scenarios, the labeled data resource is limited and contains a certain amount of noise. The fact leads the above task-balanced MTL methods to perform suboptimally and struggle to achieve promising task prediction results.

In recent years, some works \cite{DBLP:journals/corr/abs-2007-00339,DBLP:conf/ijcnn/FreitasBGM22} introduce the information bottleneck (IB) principle \cite{tishby1999information,tishby2015deep} into the information encoding process of multi-task learning to enhance the adaptability to noisy data. Different from vanilla MTL, IB-based MTL methods explicitly compress task-irrelevant redundant information by minimizing the mutual information between the input and the task-agnostic representations during the multi-task encoding process. However, in multi-task scenarios, redundant information often differs across tasks, leading to situations where information beneficial for one task may become redundant for another. For instance, target features that are highly relevant to stance detection may be irrelevant in emotion recognition. Directly applying the IB principle to compress redundancy for one task is prone to losing necessary information for other tasks. As a result, the learned shared representations would not only retain some redundant features but also face the task-specific insufficiency issue.

In this paper, we propose a new principled multi-task representation learning framework, named InfoMTL, to extract noise-invariant sufficient representations for all tasks. It ensures sufficiency of shared representations for all tasks and mitigates the negative effect of redundant features, which can enhance language understanding of pre-trained language models (PLMs) under the multi-task paradigm.

Firstly, we propose a shared information maximization (SIMax) principle to learn more sufficient shared representations for all target tasks. The SIMax principle simultaneously maximizes the mutual information between the input $X$ and the shared representations $Z$ for all tasks, as well as the mutual information between the shared representations $Z$ and the target $Y_t$ for each task $t$. It can preserve key information from the input and retain the necessary information for all target tasks during the network's implicit compression process.
In the implementation of SIMax, we utilize noise-contrastive estimation \cite{DBLP:journals/jmlr/GutmannH10,DBLP:journals/corr/abs-1807-03748} to optimize the lower bound of $I(X;Z)$ in the multi-task encoding process. As shown in Figure~\ref{fig:compare}(b) and (c), unlike IB-based MTL methods \cite{DBLP:journals/corr/abs-2007-00339, DBLP:conf/ijcnn/FreitasBGM22} that minimize the input-representation information, our InfoMTL with SIMax handles the information in an opposite manner to avoid the insufficiency issue arising from representation compression in the multi-task paradigm.

Besides, we design a task-specific information minimization (TIMin) principle to mitigate the negative effect of potential redundant features in the input for each task. The TIMin principle minimizes the mutual information between the input $X$ and the task-specific output representations $Z_t$, while maximizing the mutual information between the output representations $Z_t$ and the corresponding target task $Y_t$ for each task $t$. 
It can compress task-irrelevant redundant information and preserve the necessary information relevant to the target for multi-task prediction. In the implementation of TIMin, we perform probabilistic embedding \cite{DBLP:journals/corr/VilnisM14,DBLP:conf/aaai/0001WLZH24} and task prediction in the multi-task decoding process.
As shown in Figure~\ref{fig:compare}(b) and (c), unlike IB-based MTL methods that reduce redundancy in the task-agnostic (shared or intermediate) representations, our TIMin focuses on alleviating redundancy in the task-specific output representations to avoid redundant interference across tasks. In this way, compressing task-irrelevant information does not interfere with the sufficiency of shared representations for other tasks.

We conduct experiments on six natural language understanding benchmarks. The results show that our InfoMTL outperforms 12 representative MTL methods across different PLMs under the same multi-task architecture.
For example, with the RoBERTa backbone, 
InfoMTL improves the average performance by \textbf{+2.33\%} and achieves an average relative improvement ($\Delta p$) of \textbf{+3.97\%} over the EW baseline. Compared to single task learning baselines, InfoMTL also achieves better results on most tasks with the same scale of model parameters. Extensive experiments demonstrate that InfoMTL offers significant advantages in data-constrained and noisy scenarios, with learned representations being more sufficient, data-efficient, and robust.

The main contributions are summarized as follows:  
1) We propose a shared information maximization (SIMax) principle to learn more sufficient shared representations for all target tasks. It can avoid the insufficiency issue arising from representation compression in the multi-task paradigm.
2) We design a task-specific information minimization (TIMin) principle to mitigate the negative effect of potential redundant features in the input for each task. It can compress task-irrelevant redundant information and preserve necessary information relevant to the target for multi-task prediction.
3) We design a new principled multi-task learning framework (InfoMTL) to extract noise-invariant sufficient representations for all tasks. It can enhance language understanding of PLMs under the multi-task paradigm.
4) Experiments on six benchmarks show that our method outperforms comparative MTL approaches under the same multi-task settings, especially in data-constrained and noisy scenarios. Extensive experiments demonstrate that the learned representations are more sufficient, data-efficient, and robust.\footnote{The source code is available at \url{https://github.com/zerohd4869/InfoMTL}}

\section{Preliminary}
\paragraph{Scope of the Study.}
This paper follows the line of multi-task optimization that typically employs a hard parameter-sharing pattern \cite{DBLP:conf/icml/Caruana93}, where several lightweight task-specific heads are attached to a heavyweight task-agnostic backbone.  Another orthogonal line of MTL research focuses on designing network architectures that usually employ a soft parameter-sharing pattern. The details of these two lines are listed in the Appendix.

\paragraph{Notations.} Suppose there are $T$ tasks and task $t$ has its corresponding dataset $\mathcal{D}_t$. An MTL model typically involves two parametric modules, i.e., a shared encoder with parameters $\theta$, and $T$ task-specific decoders with parameters $\{\phi_t\}_{t=1}^{|T|}$, where $\phi_t$ represents the parameters for task $t$.
Let $\ell_t(\mathcal{D}_t;\theta,\phi_t)$ be the average loss on $\mathcal{D}_t$ for $t$. $\{\lambda_t\}_{t=1}^{|T|}$ are task-specific loss weights with a constraint that $\lambda_t\ge0$ for $t$.

\paragraph{MTL Baseline.} Since there are multiple losses in MTL, they usually are aggregated as a single one via loss weights, $\mathcal{L}(\mathcal{D};\theta, \{\phi_t\}_{t=1}^{|T|})=\sum_{t=1}^{|T|}\lambda_t\ell_t(\mathcal{D}_t;\theta,\phi_t)$. Apparently, the most simple method for loss weighting is to assign the same weight to all the tasks in the training process, i.e., $\lambda_t=\frac{1}{|T|}$ for task $t$ in every iteration. The method is a common baseline in MTL named EW in this paper. 

\paragraph{Information Flow in Neural Networks.} Let $X$ be an input random variable, $Y_t$ be a target variable given task $t$, and $p(x,y_t)$ be their joint distribution for task $t$. The universal representations $Z$ shared by all tasks is a function of $X$ by a mapping $p_\theta(z|x)$. For task $t$, the output representations $Z_t$ in the output space can be obtained by a task-specific head $p_{\phi_t}(z_t|z)$, and the corresponding prediction variable $\hat{Y}_t$ is non-parametric mapping of $Z_t$.  Then, define information flow \cite{DBLP:journals/corr/Shwartz-ZivT17,DBLP:conf/icml/GoldfeldBGMNKP19} in neural networks as a Markov chain shared by all tasks, i.e., $Y_t \rightarrow X \rightarrow Z \rightarrow Z_t \rightarrow \hat{Y}_t$ for task $t$.

\section{Methodology}
To ensure sufficiency for target tasks and mitigate the negative effects of redundant features, we propose a principled multi-task representation learning framework (InfoMTL) that extracts noise-invariant sufficient representations for all tasks. It contains two learning principles: shared information maximization (SIMax) and task-specific information minimization (TIMin), which constrain the amount of shared and task-specific information in the multi-task learning process.

\subsection{Shared Information Maximization Principle} 
In the MTL paradigm, the mutual information between the input $X$ and the shared representations $Z$ is typically reduced implicitly  (Figure~\ref{fig:compare}(a)) or explicitly (Figure~\ref{fig:compare}(b)) under the information bottleneck (IB) theory \cite{DBLP:journals/corr/Shwartz-ZivT17, DBLP:conf/icml/KawaguchiDJH23}. However, directly reducing redundancy for one task is prone to losing necessary information for others. As a result, the learned shared representations $Z$ usually suffer from task-specific insufficiency.

To alleviate the insufficiency issue, a shared information maximization (SIMax) principle is proposed to learn more sufficient shared representations for all target tasks.
The SIMax principle simultaneously maximizes the mutual information between the input $X$ and the shared representations $Z$ for all tasks, as well as the mutual information between the shared representations $Z$ and the target task $Y_t$ for each task $t$. It can be formulated as the maximization of the following Lagrangian,
\begin{equation}
    \max \sum_{t=1}^{|T|}[I(Y_t;Z)] + \alpha I(X;Z), 
    \label{eq:overall0}
\end{equation}
subject to the Markov constraint, i.e., $Y_t \rightarrow X \rightarrow Z \rightarrow Z_t  \rightarrow \hat{Y}_t$. 
$\alpha$ is a parameter that balances the trade-off between the informativeness of $Z$ for $X$ and $Y_t$.

In Equation~(\ref{eq:overall0}), the second term promotes the task-agnostic shared representations $Z$ preserves as much information as possible about the input $X$. It ensures the sufficiency of the shared representations $Z$ for all potential targets $Y$. The first term encourages the shared representations $Z$ to capture the necessary information relevant to the target $Y_t$ for each task $t$.
As shown in Figure~\ref{fig:compare}(b) and (c), unlike IB-based MTL methods \cite{DBLP:journals/corr/abs-2007-00339, DBLP:conf/ijcnn/FreitasBGM22} that explicitly minimize input-representation information, our InfoMTL with SIMax handles the information in an opposite manner to avoid the insufficiency issue arising from representation compression in the multi-task paradigm.

\paragraph{Implementation of SIMax}
The implementation of SIMax contains two terms, i.e., maximizing $\sum_{t=1}^{|T|}[I(Y_t;Z)]$ and $I(X;Z)$. 
Firstly, we maximize the lower bound of $I(Y_t;Z)$ 
by estimating the conditional entropy of the target $Y_t$ given the shared representations $Z$. Following \citet{DBLP:journals/entropy/KolchinskyTW19} and \citet{DBLP:conf/aaai/0001WLZH24}, we use cross-entropy (CE) as the estimator for each classification task $t$.
Secondly, to maximize $I(X;Z)$, we use the InfoNCE estimator \cite{DBLP:journals/jmlr/GutmannH10,DBLP:journals/corr/abs-1807-03748} to optimize the lower bound of $I(X;Z)$ during the multi-task encoding process. According to the information flow of  $Z \leftarrow X \rightarrow Z^\prime$, the Markov chain rule states that $I(X; Z) \geq I(Z; Z^\prime)$. For the InfoNCE estimation of maximizing $I(Z; Z^\prime)$, the selection of positive and negative samples as well as the implementation of noise-contrastive loss are consistent with \citet{DBLP:conf/acl/0001WZH24}. The optimization objective of SIMax for MTL can be:
\begin{equation}
\resizebox{1.0\linewidth}{!}{$
\begin{aligned}
    \mathcal{L}_{\text{SIMax}} = \mathbb{E}_{z \sim {p_{\theta}(z|x)}} \{ & \mathbb{E}_{t \sim T} \left[-\log q_{\phi_t}(y_{t}|z)\right] \\
    & - \alpha \log \frac{\exp(\text{sim}(z, z^{+}) / \tau)}{\sum_{z' \in \mathcal{B^{+}}} \exp(\text{sim}(z, z') / \tau)}  \}, 
\end{aligned}
$}
\end{equation}
where 
$p_{\theta}(z|x)$ is a shared encoder with parameters $\theta$.
$q_{\phi_t}(y_{t} | z)$ is a task-specific decoder with parameters ${\phi_t}$ for task $t$, and its output distribution is adapted for task prediction by a non-parametric function (e.g., Softmax operation for classification). $z^+$ refers to the positive key of $z$, generated by dropout in $p_{\theta}(z|x)$. $\mathcal{B^{+}}$ represents the set of positive keys in the current batch $\mathcal{B}$. $\text{sim}(\cdot)$ is a pairwise similarity function, i.e., cosine similarity. $\tau > 0$ is a scalar temperature parameter that controls the sharpness of the probability distribution, which is applied during the Softmax operation.

\subsection{Task-specific Information Minimization Principle}
In multi-task scenarios, redundant information often differs across tasks, such that information beneficial for one task may become redundant for another. Directly applying the IB principle \cite{tishby1999information,tishby2015deep} to compress the task-specific redundancy for one task can result in the loss of necessary information for other tasks.
Consequently, both the shared representations $Z$ and the task-specific output representations $Z_t$ for a given task $t$ often contain redundant features that are irrelevant to the specific task.

To better alleviate the redundancy issue, a task-specific information minimization (TIMin) principle is designed to mitigate the negative effect of potential redundant features in the inputs for the target task. 
It can compress task-irrelevant redundant information and preserve necessary information relevant to the target for multi-task prediction.
The TIMin principle minimizes the mutual information between the input $X$ and the task-specific output representations $Z_t$, while maximizing the mutual information between the output representations $Z_t$ and the corresponding target task $Y_t$ for each task $t$.
The principle can be formulated as the maximization of the following Lagrangian,
\begin{equation}
    \small
    \max \sum_{t=1}^{|T|} [I(Y_t;Z_t) - \beta I(X;Z_t)], 
    \label{eq:overall1}
\end{equation}
subject to $Y_t \rightarrow X \rightarrow Z \rightarrow Z_t  \rightarrow \hat{Y}_t$ for each task $t$. 
$\beta$ is a trade-off parameter of the compression of $Z_t$ from the input $X$ and the informativeness of $Z_t$ for ${Y_t}$.

In Equation~(\ref{eq:overall1}), the first term encourages the output representations $Z_t$ to preserve task-relevant information necessary for multi-task prediction. The second term compresses task-irrelevant redundant information of $Z_t$ for each task. 
These two terms make the learned output representations $Z_t$ approximately the minimal sufficient task-specific representations for each task $t$.
As shown in Figure~\ref{fig:compare}(b) and (c), unlike IB-based MTL methods that reduce redundancy in the task-agnostic (shared or intermediate) representations, our TIMin focuses on alleviating redundancy in the task-specific output representations to avoid redundant interference across tasks. In this way, compressing task-irrelevant information does not interfere with the sufficiency of shared representations for other tasks.

\paragraph{Implementation of TIMin}
To achieve the principle of TIMin, we perform probabilistic embedding \cite{DBLP:journals/corr/VilnisM14,DBLP:conf/aaai/0001WLZH24} and task prediction in the multi-task decoding process.
Following \citet{DBLP:conf/aaai/0001WLZH24}, we perform variational inference \cite{DBLP:journals/jmlr/HoffmanBWP13} to minimize the mutual information between the input $X$ and task-specific output representations $Z_t$ for each task $t$. 
It maps the shared representations $Z$ to a set of different Gaussian distributions in the output space, i.e., $\mathbb{R}^{|\mathcal{Y}_t|}$.
Additionally, we can maximize the lower bound of $I(Y_t;Z_t)$ by estimating the conditional entropy $H(Y_t|Z_t)$.

Given the input $x$ and its task-agnostic representations $z$, the task-specific output representations $z_t \sim p_{\theta,\phi_t}(z_t|x)$ can be learned by the shared encoder with parameters $\theta$ and task-specific  head with parameters $\phi_t$. The true posterior $p_{\theta,\phi_t}(z_t|x)$ can be approximated as $p_{\phi_t}(z_t|z)$ where $z \sim {p_{\theta}}(z|x)$. 
Let $r(z_t) \sim \mathcal{N}(z_t; \mathbf{0}, \mathbf{I})$ be an estimate of the prior $p(z_t)$ of $z_t$.
Let $q_{\phi_t}(z_t|z)$ be a variational estimate of the intractable true posterior $p(z_t|z)$ of $z_t$ given $z$, and learned by the $t$-th stochastic head parametrized by $\phi_t$.
Then, we have $I(Z;Z_t)= \int dz \, dz_t \, p(z,z_t) \log \frac{p(z_t|z)}{p(z_t)} \lesssim \int dz \, dz_t \, p(z) \, q(z_t|z) \log \frac{q(z_t|z)}{r(z_t)} $.
And the optimization objective of TIMin for MTL can be:
\begin{equation} 
\resizebox{1.0\linewidth}{!}{$
\begin{split}
      \mathcal{L}_{\text{TIMin}} = \mathbb{E}_{t \sim T, z \sim {p_{\theta}(z|x)}} \{ & \mathbb{E}_{z_t \sim {q_{\phi_t}(z_t|z)}} [-\log s(y_{t}|z_{t})] \\
      & + \beta  KL(q_{\phi_t}(z_t|z); r(z_t)) \},
\end{split}
$}
\end{equation}
where $z_t$ is randomly sampled from $p_{\phi_t}(z_t|z)$ and $s(y_{t}|z_t)$ is a non-parametric operation on $z_t$. $KL(\cdot)$ denotes the analytic KL-divergence term, serving as the regularization that forces the variational posterior $q_{\phi_t}(z_t|z)$ to approximately converge to the Gaussian prior $r(z_t)$. 
$\beta > 0$ controls the trade-off between the sufficiency of $z_t$ for predicting $y_t$, and the compression of $z_t$ from $x$.

We assume the variational posterior $q_{\phi_t}(z_t|z)$ be a multivariate Gaussian with a diagonal covariance structure, i.e., 
\begin{equation}
q_{\phi_t}(z_t^i|z^i) = \mathcal{N}(z^i_t; \mu_t(z^i), \Sigma_t(z^i)), \label{eq:norm}
\end{equation}
where $\mu_t(z^i)$ and $\Sigma_t(z^i)$ denote the mean and diagonal covariance of sample $z^i$ for task $t$. 
Following \citet{DBLP:conf/emnlp/0001HDZJMS22,DBLP:conf/aaai/0001WLZH24}, both of their parameters are input-dependent and can be learned by an MLP (a fully-connected neural network with a single hidden layer) for each task, respectively.
Next, we sample $z_t^i$ from the approximate posterior $q_{\phi_t}(z_t^i|z^i)$, and obtain the prediction value by $s(y_t^i|z_t^i)$.
Since the sampling process of $z_t^i$ is stochastic, we use the re-parameterization trick \cite{DBLP:journals/corr/KingmaW13} to ensure it trainable, i.e.,
$z_t^i =  {\mu}_t(z^i) +  {({\Sigma}_t(z^i))}^{1/2}  \odot \epsilon,  \epsilon \sim \mathcal{N}(\mathbf{0}, \mathbf{I}),$
where $\odot$ refers to an element-wise product.
Then, the KL term can be calculated by:
${KL}(q_{\phi_t}({z}^i_t  | {z}^i);r({z}^i_t))  =   -\frac{1}{2} \left( 1 + \log {\Sigma}_t(z^i) - ({\mu}_t(z^i))^2  - {\Sigma}_t(z^i) \right)$.

\subsection{InfoMTL Framework}
We incorporate the TIMin principle into the SIMax principle, and design a new information-theoretic multi-task learning framework (InfoMTL) to extract noise-invariant sufficient representations for all tasks.
According to data processing inequality, in the Markov chain $Y_t \rightarrow X \rightarrow Z \rightarrow Z_t$, we have $I(Y_t;Z) \geq I(Y_t;Z_t)$. 
For simplification, we use $\max I(Y_t;Z_t)$ in TIMin to compute the lower bound of $\max I(Y_t;Z)$ in SIMax.
As shown in Figure~\ref{fig:compare}(c), the total learning principle of InfoMTL is,
\begin{equation}
\begin{split}
    \max \sum_{t=1}^{|T|}[I(Y_t;Z_t)- \beta I(X;Z_t)] + \alpha I(X;Z). 
    \label{eq:overall2}
\end{split}
\end{equation}
Firstly, maximizing $I(X;Z)$ can learn more sufficient shared representations for all target tasks. It preserves as much information as possible about the input $X$ and ensures sufficiency of the task-agnostic shared representations for all targets $Y$. 
Then, minimizing $I(X;Z_t)$ mitigates the negative effect of potential redundant features in the input for each task. 
It can compress task-irrelevant redundant information.
Finally, maximizing $I(Y_t;Z_t)$ captures necessary information relevant to the target $Y_t$ from the output representations $Z_t$. 
It ensures sufficiency of the task-agnostic shared representations and task-specific output representations for multi-task prediction.
Totally, InfoMTL can preserve necessary information in the shared representations for all tasks, and eliminate redundant information in the task-specific representations for each task.

\section{Experiments}
\subsection{Experimental Setups}
\paragraph{Datasets and Downstream Tasks}
Since this paper mainly focuses on MTL in natural language understanding,  we experiment on six text classification benchmarks \cite{DBLP:conf/emnlp/BarbieriCAN20}, i.e., \textit{EmotionEval} \cite{DBLP:conf/semeval/MohammadBSK18} for emotion detection, \textit{HatEval} \cite{DBLP:conf/semeval/BasileBFNPPRS19} for hate speech detection, \textit{IronyEval} \cite{DBLP:conf/semeval/HeeLH18} for irony detection, \textit{OffensEval} \cite{DBLP:conf/semeval/ZampieriMNRFK19} for offensive language detection, \textit{SentiEval} \cite{DBLP:conf/semeval/RosenthalFN17} for sentiment analysis, and \textit{StanceEval} \cite{DBLP:conf/semeval/MohammadKSZC16} for stance detection.
The details are listed in the Appendix.

\paragraph{Comparison Methods}
To fairly compare our method with different multi-task methods, we reproduce and compare with the following 12 representative MTL methods under the same experimental settings (e.g., network architecture).
Comparison methods include Equal Weighting (EW), {Task Weighting} (TW), {Scale-invariant Loss} (SI), {Uncertainty Weighting} (UW) \cite{DBLP:conf/cvpr/KendallGC18}, {Geometric Loss Strategy} (GLS) \cite{DBLP:conf/cvpr/ChennupatiSYR19},
{Dynamic Weight Average} (DWA) \cite{DBLP:conf/cvpr/LiuJD19},
{Projecting Conflicting Gradient} (PCGrad)  \cite{DBLP:conf/nips/YuK0LHF20}, 
{IMTL-L} \cite{DBLP:conf/iclr/LiuLKXCYLZ21},
{Random Loss Weighting} (RLW) \cite{DBLP:journals/tmlr/LinYZT22}, {MT-VIB} \cite{DBLP:journals/corr/abs-2007-00339},
{VMTL} \cite{DBLP:conf/nips/ShenZWS21},
and {Hierarchical MTL}  \cite{DBLP:conf/ijcnn/FreitasBGM22}.
MT-VIB, VMTL, Hierarchical MTL are probabilistic MTL series.
We use two pre-trained language models, i.e., BERT \cite{DBLP:conf/naacl/DevlinCLT19} and RoBERTa \cite{DBLP:journals/corr/abs-1907-11692}, as the backbone model.
Concretely, we use \textit{bert-base-uncased}\footnote{\url{https://huggingface.co/}\label{code}} and \textit{roberta-base}\textsuperscript{\ref{code}} to initialize BERT and RoBERTa for fine-tuning.
For each method, we fine-tune key parameters following the original paper to obtain optimal performance.
We also compare with the single task learning baseline (STL) and the large language model GPT-3.5\footnote{\url{https://chat.openai.com}}.
See the Appendix for more details.

\begin{table}[t]
\centering
\small
\begin{tabular}{l|cccc}
\hline
\multicolumn{1}{c|}{\multirow{2}{*}{{Methods}}} & \multicolumn{2}{c}{\textit{BERT backbone}} & \multicolumn{2}{c}{\textit{RoBERTa backbone}} \\  
& \multicolumn{1}{c}{\multirow{1}{*}{\textbf{Avg.}}} 
& \multicolumn{1}{c}{\multirow{1}{*}{$\mathbf{\Delta p} \uparrow$}} 
& \multicolumn{1}{c}{\multirow{1}{*}{\textbf{Avg.}}} 
& \multicolumn{1}{c}{\multirow{1}{*}{$\mathbf{\Delta p} \uparrow$}} 
\\ \hline 
EW (baseline)  & 65.62 & 0.00  & {66.17} & 0.00 \\
\hdashline
\multicolumn{5}{l}{\multirow{1}{*}{\textit{Task-balanced Methods}}} \\ 
SI  & 65.67 & +0.06  &  67.16 & +1.75 \\
TW  & 65.68 & +0.11  & 67.08 & +1.55 \\
UW & 66.97 & +2.22 & 67.11 & +1.92 \\ 
GLS   & 66.05 & +0.60 & 67.32 & +1.67 \\ 
DWA   & 65.56 & -0.09 & {66.94} & +1.35 \\
PCGrad   & 65.45 & -0.50 & 67.42 & +1.96 \\ 
IMTL-L   & 66.18 & +0.86 & 66.54  & +0.67 \\ 
RLW   & 66.76 & +1.86 & 67.07 &  +1.63 \\
\multicolumn{5}{l}{\multirow{1}{*}{\textit{Probabilistic Methods}}} \\
MT-VIB  & 65.80 & +0.66 & 67.14 & +2.00  \\ 
VMTL  & 65.80 & +0.65 & 67.05 & +1.81 \\
Hierarchical MTL   & 66.42 & +1.76 & 66.84 & +1.60 \\
\textbf{InfoMTL} (ours) &  
\textbf{67.51}$^{*}$ & \textbf{+3.70}
& \textbf{68.50}$^{*}$ & \textbf{+3.97} \\
\hline 
\end{tabular}
\caption{Multi-task performance (\%) on six benchmarks.  
For all methods with BERT/RoBERTa backbone, we run three random seeds and report the average result on test sets.
Best results are highlighted in bold. 
$^{*}$ represents statistical significance over scores of the baseline under the $t$-test ($p < 0.05$).}
\label{tab:overall_results}
\end{table}

\begin{table*}[t]
\centering
\small
\resizebox{0.99\linewidth}{!}{$
\begin{tabular}{l|cccccc|cc}
\hline
\multicolumn{1}{c|}{\multirow{2}{*}{{Methods}}} 
& \multicolumn{1}{c}{{EmotionEval}} 
& \multicolumn{1}{c}{{HatEval}} 
& \multicolumn{1}{c}{{IronyEval}} 
& \multicolumn{1}{c}{{OffensEval}} 
& \multicolumn{1}{c}{{SentiEval}} 
& \multicolumn{1}{c|}{{StanceEval}} 
& \multicolumn{1}{c}{\multirow{2}{*}{\textbf{Avg.}}} 
& \multicolumn{1}{c}{\multirow{2}{*}{$\mathbf{\Delta p} \uparrow$}} 
\\  
& M-F1 & M-F1 &  F1(i.) & M-F1 & M-Recall & M-F1 (a. \& f.) & \\
\hline 
EW (baseline)  
& {74.37}\scriptsize{$\pm$0.56} & {44.08}\scriptsize{$\pm$5.26} & {65.32}\scriptsize{$\pm$1.84} & {79.04}\scriptsize{$\pm$1.43} & {70.64}\scriptsize{$\pm$1.71} & {63.59}\scriptsize{$\pm$2.43} & {66.17}\scriptsize{$\pm$0.43} & 0.00 \\
\hdashline
MT-VIB 
&74.74\scriptsize{$\pm$0.38} & 48.06\scriptsize{$\pm$4.79} & 66.09\scriptsize{$\pm$3.38} & 78.17\scriptsize{$\pm$1.39} & 70.95\scriptsize{$\pm$0.99} & 64.83\scriptsize{$\pm$1.56} & 67.14\scriptsize{$\pm$0.87} & +2.00 \\ 
VMTL 
& 74.07\scriptsize{$\pm$0.72} & 47.44\scriptsize{$\pm$3.42}  & {68.55}\scriptsize{$\pm$2.80} & 77.95\scriptsize{$\pm$0.22} & 70.52\scriptsize{$\pm$1.04} & 63.76\scriptsize{$\pm$2.86} & 67.05\scriptsize{$\pm$1.06} & +1.81 \\
Hierarchical MTL 
& 74.09\scriptsize{$\pm$1.73} & \textbf{48.52}\scriptsize{$\pm$4.26}  & 64.92\scriptsize{$\pm$6.14} & 78.26\scriptsize{$\pm$1.63} & {71.45}\scriptsize{$\pm$0.44} & 63.82\scriptsize{$\pm$0.54} & 66.84\scriptsize{$\pm$1.68} & +1.60 \\ 
\textbf{InfoMTL} (ours) & \textbf{76.90}$^{*}$\scriptsize{$\pm$0.62} & {48.44}$^{*}$\scriptsize{$\pm$2.15} & \textbf{68.94}$^{*}$\scriptsize{$\pm$1.86} & \textbf{79.78}$^{*}$\scriptsize{$\pm$0.86}  & \textbf{71.92}$^{*}$\scriptsize{$\pm$0.36} & \textbf{65.02}$^{*}$\scriptsize{$\pm$1.81} & \textbf{68.50}$^{*}$\scriptsize{$\pm$0.58} & \textbf{+3.97} \\
\hline
\end{tabular}
$}
\caption{Fin-grained results (\%) of probabilistic multi-task methods with RoBERTa backbone.  
$^{*}$ represents statistical significance over scores of the baseline under the $t$-test ($p < 0.05$).
}
\label{tab:fine_results}
\end{table*}

\begin{table*}[t]
\centering
\small
\begin{tabular}{l|c|cccccc|c}
\hline
\multicolumn{1}{c|}{\multirow{2}{*}{{Methods}}} 
& \multicolumn{1}{c|}{\multirow{2}{*}{\# Param.}} 
& \multicolumn{1}{c}{{EmotionEval}} 
& \multicolumn{1}{c}{{HatEval}} 
& \multicolumn{1}{c}{{IronyEval}} 
& \multicolumn{1}{c}{{OffensEval}} 
& \multicolumn{1}{c}{{SentiEval}} 
& \multicolumn{1}{c|}{{StanceEval}} 
& \multicolumn{1}{c}{\multirow{2}{*}{\textbf{Avg.}}} 
\\  
& & M-F1 & M-F1 &  F1(i.) & M-F1 & M-Recall & M-F1 (a. \& f.)  \\
\hline 
GPT-3.5 & (LLMs) & {73.23} &  {48.30} & {66.81}  & {63.71}  & {40.40} & {39.45} & 55.32    \\ \hdashline
STL with CNN & 110M+6$\times$2M & 59.11 & 47.61 & 52.10 & 77.80 & 70.85 & 57.58 & 60.84 \\
\textbf{InfoMTL} & 110M & \textbf{76.90} & \textbf{48.44} & \textbf{68.94} & \textbf{79.78}   & \textbf{71.92}  & \textbf{65.02} & \textbf{68.50}  \\ 
\hline 
STL & 6$\times$110M & {74.49}  & 45.26  & 53.27 & {79.20} & \textbf{72.43} & 66.70 & 65.23  \\ 
\textbf{InfoMTL}* & $\textless$ 6${\times}${110M}  & \textbf{78.55} & \textbf{52.38} & \textbf{68.14} & \textbf{80.80}   & {72.40}  & \textbf{68.56} & \textbf{70.14}   \\
\hline
\end{tabular}
\caption{Comparison results (\%) with different learning paradigms.
We experiment with RoBERTa backbone for all methods except for GPT-3.5.
STL means single-task learning with a cross-entropy loss. 
STL with CNN indicates fine-tuning task-specific CNN classifiers with a frozen RoBERTa backbone. 
InfoMTL and InfoMTL* indicate the model trained on six and pair-wise tasks, respectively.
\# Param. refers to the number of model parameters for all tasks excluding the task-specific linear head.}
\label{tab:single}
\end{table*}

\begin{table}[t]
\centering
\small
\resizebox{0.99\linewidth}{!}{$
\begin{tabular}{l|cccc}
\hline
\multicolumn{1}{c|}{\multirow{2}{*}{{Methods}}} 
& \multicolumn{2}{c}{\textit{BERT backbone}}  &  \multicolumn{2}{c}{\textit{RoBERTa backbone}}  \\ 
& \multicolumn{1}{c}{\textbf{Avg.}} 
& \multicolumn{1}{c}{$\mathbf{\Delta p} \uparrow$}
& \multicolumn{1}{c}{\textbf{Avg.}} 
& \multicolumn{1}{c}{$\mathbf{\Delta p} \uparrow$}
\\  
\hline
\textbf{InfoMTL} 
    & \textbf{67.51} & \textbf{+3.70}  
    & \textbf{68.50} & \textbf{+3.97} \\  
\ \ w/o SIMax   
    & 66.01 & +0.71 
    & 67.59 & +2.72 \\    
\ \ w/o TIMin   
    & 67.02 & +2.59
    & 67.55 & +2.34 \\  
 \ \ w/o SIMax \& TIMin 
    & 65.62 & 0.00
    & {66.17} & 0.00 
 \\
\hline
\end{tabular}
$}
\caption{Ablation results (\%) of our InfoMTL.}
\label{tab:ablated}
\end{table}

\paragraph{Evaluation Metrics}
We use the same evaluation metric as the original tasks. The macro-averaged F1 over all classes is applied in EmotionEval, HatEval, and OffensEval. 
The F1 score of ironic class is applied in IronyEval. The macro-averaged F1 score of favor and against classes is applied in StanceEval.
The macro-averaged recall score is applied in SentiEval. 
Following \citet{DBLP:conf/emnlp/BarbieriCAN20,DBLP:conf/aaai/0001WLZH24}, we report a global metric based on the average of all task-specific metrics, denoted as $\mathbf{Avg.}$. 
Following \citet{DBLP:conf/cvpr/ManinisRK19,DBLP:journals/tmlr/LinYZT22}, we also report the average relative improvement over the EW baseline, denoted as $\mathbf{\Delta}p$. 
In addition, the $t$-test \cite{kim2015t} is used to verify the statistical significance of the differences between the results of our method and the baseline.

\paragraph{Implementation Details}
All experiments are conducted on a single NVIDIA Tesla A100 80GB card. 
The validation sets are used to tune hyperparameters and choose the optimal model.
For each method, we run three random seeds and report the average result of the test sets.
Besides, 
we experiment using an epoch number of $20$, a total batch size of $128$, and a maximum token length of $256$. The maximum patience for early stopping is set to $3$ epochs. 
The network parameters are optimized by using Adamax optimizer \citep{DBLP:journals/corr/KingmaB14}.
The dropout is searched from $\{0, 0.2\}$.
The parameters $\alpha$ and $\beta$ are searched from $\{0.001, 0.01, 0.1, 1\}$.
$\tau$ is searched from $\{0.1, 1\}$.
See the Appendix for more details.

\subsection{Main Results}
\paragraph{Overall Results for MTL}
The overall results on six benchmarks with the BERT and RoBERTa backbone are summarized in Table~\ref{tab:overall_results}. 
For each backbone, the top row shows the performance of the widely used EW, and we use it as a baseline to measure the relative improvement of different methods as shown in the definition of $\Delta p$.
From the results, our InfoMTL achieves the best performance in terms of Avg and $\Delta p$ on different backbones. 
With the BERT/RoBERTa backbone, InfoMTL enhances the average performance by \textbf{+1.89\%}/\textbf{+2.33\%} and achieves $\Delta p$ of \textbf{+3.70\%}/\textbf{+3.97\%} over the EW baseline.

\paragraph{Fine-grained Results for Probabilistic MTL} 
Table~\ref{tab:fine_results} shows the comparison of InfoMTL and representative probabilistic MTL methods such as MT-VIB, VMTL, and Hierarchical MTL. 
Our InfoMTL consistently outperforms EW on all tasks and achieves the best fine-grained results on most tasks, which confirms the effectiveness of our method.

\paragraph{Comparison with STL and LLM}
We compare our InfoMTL with the single-task learning (STL) baseline and the large language model (LLM) GPT-3.5. For STL, each task is trained with a separate model. For GPT-3.5, predictions are made under the zero-shot setting using task descriptions and instructions.
As shown in Table~\ref{tab:single}, our InfoMTL outperforms GPT-3.5 on all tasks significantly.
Compared to the STL baselines, our method also achieves better results on most tasks with the same scale of model parameters.

\subsection{Ablation Study}
We conduct ablation studies by removing the loss of shared information maximization ({w/o SIMax}) and task-specific information minimization ({w/o TIMin}) in our InfoMTL.
As shown in Table~\ref{tab:ablated}, the full InfoMTL achieves the best results in terms of the average performance and $\Delta p$. See the Appendix for the fine-grained results with RoBERTa backbone.
When removing either SIMax or TIMin, the ablated methods obtain inferior performance on most tasks.
When further removing both components, the ablation {w/o SIMax \& TIMin} would be equivalent to EW. 
The declining performance reveals the effectiveness of both principles.

\begin{figure}[t]
    \centering
    \includegraphics[width=0.96\linewidth]{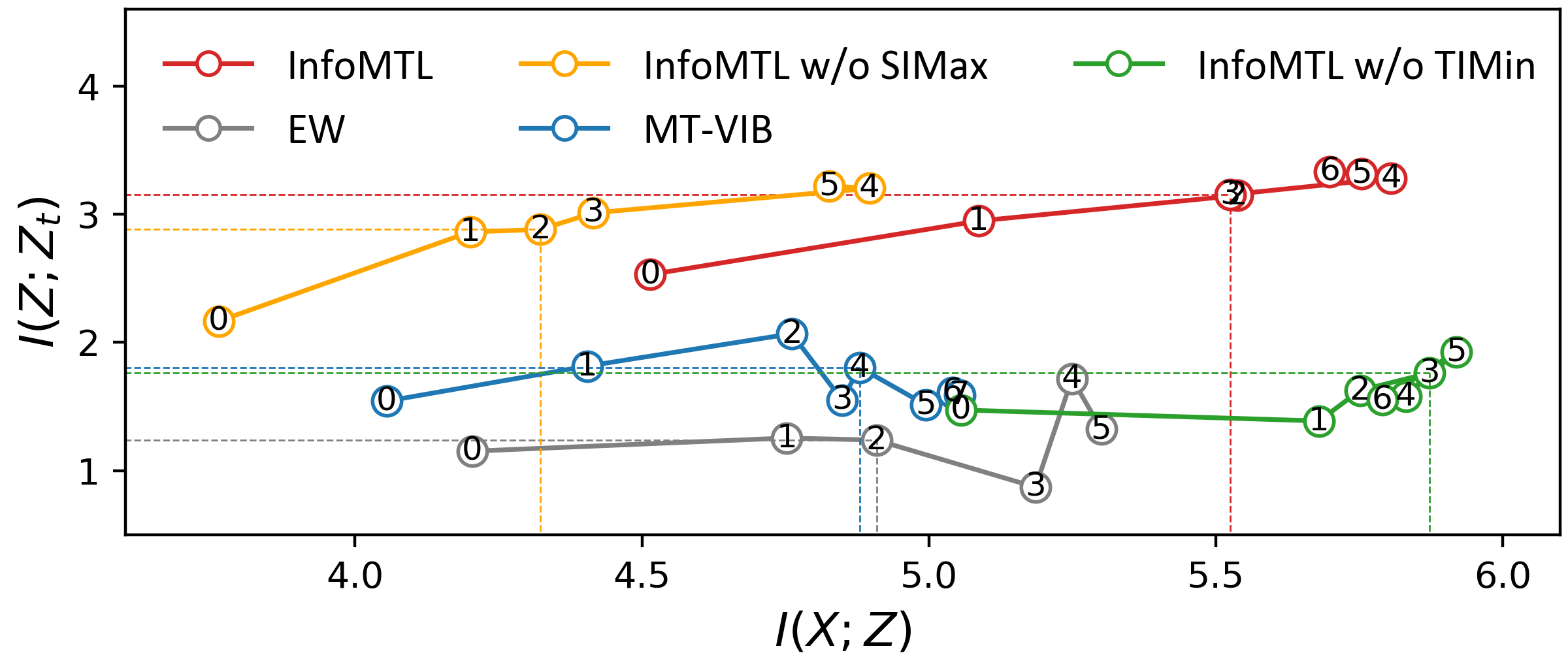}
    \caption{Mutual information analysis results. The X-axis refers to the mutual information between the shared representations $Z$ and the input $X$, i.e, $I(X;Z)$. 
    Y-axis represents the mutual information between the shared and output representations, i.e., $I(Z;Z_t)$. 
    Each number on the line is the training epoch, and  the optimal epochs are marked with dashed lines.}
    \label{fig:mi}
\end{figure}

\begin{table*}[t]
    \centering
\small
\begin{tabular}{l|cccccccccccc}
    \hline 
        \multicolumn{1}{c|}{\multirow{2}{*}{Methods}}
& \multicolumn{2}{c}{{EmotionEval}} 
& \multicolumn{2}{c}{{HatEval}} 
& \multicolumn{2}{c}{{IronyEval}} 
& \multicolumn{2}{c}{{OffensEval}} 
& \multicolumn{2}{c}{{SentiEval}} 
& \multicolumn{2}{c}{{StanceEval}}  \\ 
& ARI$\uparrow$  & Uni.$\downarrow$ 
& ARI$\uparrow$  & Uni.$\downarrow$ 
& ARI$\uparrow$  & Uni.$\downarrow$ 
& ARI$\uparrow$  & Uni.$\downarrow$ 
& ARI$\uparrow$  & Uni.$\downarrow$ 
& ARI$\uparrow$  & Uni.$\downarrow$ \\
    \hline 
EW      & 5.23 & -2.30 
        & -0.32 &-1.54  
        & 1.06 & -2.18  
        & 5.54 & -2.19 
        & 6.76  & -2.15 
        & 0.08 &   -2.35  
        \\  
UW & 8.89 & -2.21 & -0.82 & -1.54 & 0.68 & -2.21 & 0.87 & -2.18 & 4.73 & -2.13 & 1.60 & -2.21 \\ 
PCGrad & 7.21 & -2.36 & -0.03 & -1.60 & -0.28 & -2.35 & 4.02 & -2.29 & 2.97 & -2.29 & 2.13 & -2.48 \\
MT-VIB  &  1.98 & -2.25  
        & {0.14} & -1.60 
        & -0.09 & -2.26  
        & -0.17 & -2.21  
        & 1.54 & -2.13  
        &  0.01 & -2.44  \\
Hierarchical MTL 
        & 5.89 & -2.35 
        & -0.01 & -1.60
        & 0.04 & -2.38 
        & 1.13 & -2.32 
        & 0.55 & -2.21 
        & -0.00 & -2.66 \\ 
\textbf{InfoMTL} 
        & \textbf{51.38} & \textbf{-2.54}   
        & \textbf{0.29} & \textbf{-1.83}  
        & \textbf{11.40} & \textbf{-2.65}  
        & \textbf{41.58} & \textbf{-2.69}  
        & \textbf{28.66} & \textbf{-2.71} 
        & \textbf{17.04} & \textbf{-2.71} 
        \\
\hline    
\end{tabular}

\caption{Quality evaluation of the learned representations by different MTL methods.
Adjusted Rand Index (ARI, \%) assesses preserved maximal information of output representations for label structure. Uniformity (Uni.) measures preserved maximal information of hidden representations from input. The lower uniformity means the better sufficiency for the input data, and the higher ARI means the better sufficiency for target task. RoBERTa is the default backbone.}
    \label{tab:quality}
 \end{table*}

\subsection{Representation Evaluation and Analysis}
\paragraph{Mutual Information Analysis}
We analyze the mutual information between different variables in the information flow during training. 
From Figure~\ref{fig:mi}, 1) compared to the EW baseline and MT-VIB, both InfoMTL and the ablation without TIMin gain larger mutual information between $X$ and $Z$, given the same epoch. 
This indicates SIMax principle can promote the shared representations $Z$ to preserve more information about the input $X$. 
2) Compared to the EW baseline, InfoMTL without SIMax obtains less information $Z$ from $X$, and larger information $Z_t$ from $Z$, given the same epoch. 
The same trends can be observed for InfoMTL when compared to InfoMTL without TIMin. 
This indicates TIMin principle can compress the redundant features in the shared representations $Z$, and ensure sufficiency of $Z_t$ for task $t$.

\paragraph{Representation Quality Evaluation}
To evaluate the quality of representations learned by different MTL methods, we measure the sufficiency of the learned representations on the test set for both the input data and the target task.
Following \citet{DBLP:conf/acl/0001WZH24}, 
we use the uniformity (Uni.) metric \cite{DBLP:conf/icml/0001I20} to measure the preserved maximal information of the shared representations from the input, and the adjusted rand index (ARI) score to assess the preserved maximal information of output representations for label structure.
Table~\ref{tab:quality} shows Uni. and ARI of the representations learned by different MTL methods on all benchmarks.
Our InfoMTL achieves better performance on both metrics across all tasks.
This implies that InfoMTL can learn both sufficient shared representations for the input and sufficient task-specific output representations for the target task.

\subsection{Evaluation under Data-constrained Conditions}
We experiment under different ratios of the training set to evaluate generalization when training with limited data.
Following \citet{DBLP:conf/aaai/0001WLZH24,DBLP:conf/acl/0001WZH24}, all methods are trained on randomly sampled subsets from the original training set with different seeds, and we report the average results on the test set.
Table~\ref{tab:data_per} shows results of different techniques against different sizes of training set.
InfoMTL achieves superior performance against different ratios of the training set under most settings.
With only 20\% training data, InfoMTL achieves $\Delta p$ of \textbf{+4.07\%} over the EW baseline, showing better generalization of InfoMTL under data-constrained conditions.
This indicates InfoMTL can learn more sufficient representations from the inputs and enhance the efficiency of utilizing limited data.

\begin{table}[t]
\centering
\small
\begin{tabular}{l|c|cc}
\hline
\multicolumn{1}{c|}{\multirow{1}{*}{Methods}} & \multirow{1}{*}{Data per}  & \multicolumn{1}{c}{\textbf{Avg.}} & \multicolumn{1}{c}{{$\mathbf{\Delta p}$}$\uparrow$} 
\\ 
\hline
EW & 20\% & 62.43 & 0.00 \\
\hdashline
UW & 20\% & 61.78 & -1.59 \\ 
PCGrad & 20\% & 62.75 & +1.48 \\ 
MT-VIB & 20\% & 60.00 & -4.18 \\
Hierarchical MTL & 20\% & 61.11 & -2.30 \\
\textbf{InfoMTL} & 20\% & \textbf{64.83} & \textbf{+4.07} \\
\hline
EW & 40\% & 66.01  & 0.00 \\ 
\hdashline
UW & 40\% & 64.35 & -2.82 \\ 
PCGrad & 40\% & 64.28 & -2.93 \\ 
MT-VIB & 40\% & 63.58  & -3.90 \\ 
Hierarchical MTL & 40\% & 62.86  & -5.07 \\ 
\textbf{InfoMTL} & 40\% & \textbf{67.10}  & \textbf{+2.16} \\ 
\hline
EW & 60\% & 66.38  & 0.00 \\ \hdashline
UW & 60\% & 66.17 & -0.45 \\ 
PCGrad & 60\% & 65.82   & -1.21 \\
MT-VIB & 60\% & 66.31  & {+0.04} \\
Hierarchical MTL&  60\% & 65.00  & -1.95 \\ 
\textbf{InfoMTL} & 60\% & \textbf{66.71}  & \textbf{+0.50} \\  \hline
EW & 80\% & 66.34  & 0.00 \\  \hdashline
UW & 80\% & 66.93 & +1.30 \\ 
PCGrad  & 80\% & 66.48 & +0.81 \\ 
MT-VIB   & 80\% & 65.34  & -1.57 \\ 
Hierarchical MTL & 80\% & 65.35  & -1.33 \\ 
\textbf{InfoMTL} & 80\% & \textbf{67.88}  & \textbf{+2.53} \\ \hline 
\end{tabular}
\caption{Results (\%) against different sizes of training set. RoBERTa is the default backbone.}
\label{tab:data_per}
\end{table}

\subsection{Robustness Evaluation on Noisy Data} \label{sec:robust}
To assess the adaptability to noisy data \cite{fang2021animc,DBLP:conf/acl/0001WZH24}, we evaluate the model's robustness under various optimization objectives during multi-task learning. 
We adjust different strengths of random and adversarial perturbations on the test set. 
The random perturbations are from a multivariate Gaussian, and the adversarial perturbations are produced by a fast gradient method \cite{DBLP:conf/iclr/MiyatoDG17}.
These perturbations are scaled by the $L_2$ norm and then applied to the embedding layer in the testing process.
Following \citet{DBLP:conf/sp/Carlini017,DBLP:conf/acl/0001WZH24}, we report the robust scores in terms of original evaluation metrics on noise samples generated from original test sets for each task.
From Figure~\ref{fig:fgm_attack} (see the Appendix for results against different random perturbation strengths), InfoMTL gains better robust scores over other objectives on all tasks. 
Compared to EW, InfoMTL achieves an average increase of \textbf{+29.2\%} and \textbf{+27.8\%} in robust scores under random and adversarial noise, respectively.
This indicates InfoMTL extracts noise-invariant representations for all tasks, which can enhance the model's adaptability to noisy data.

\begin{figure}[t]
    \centering
        \centering
\includegraphics[width=0.99\linewidth]{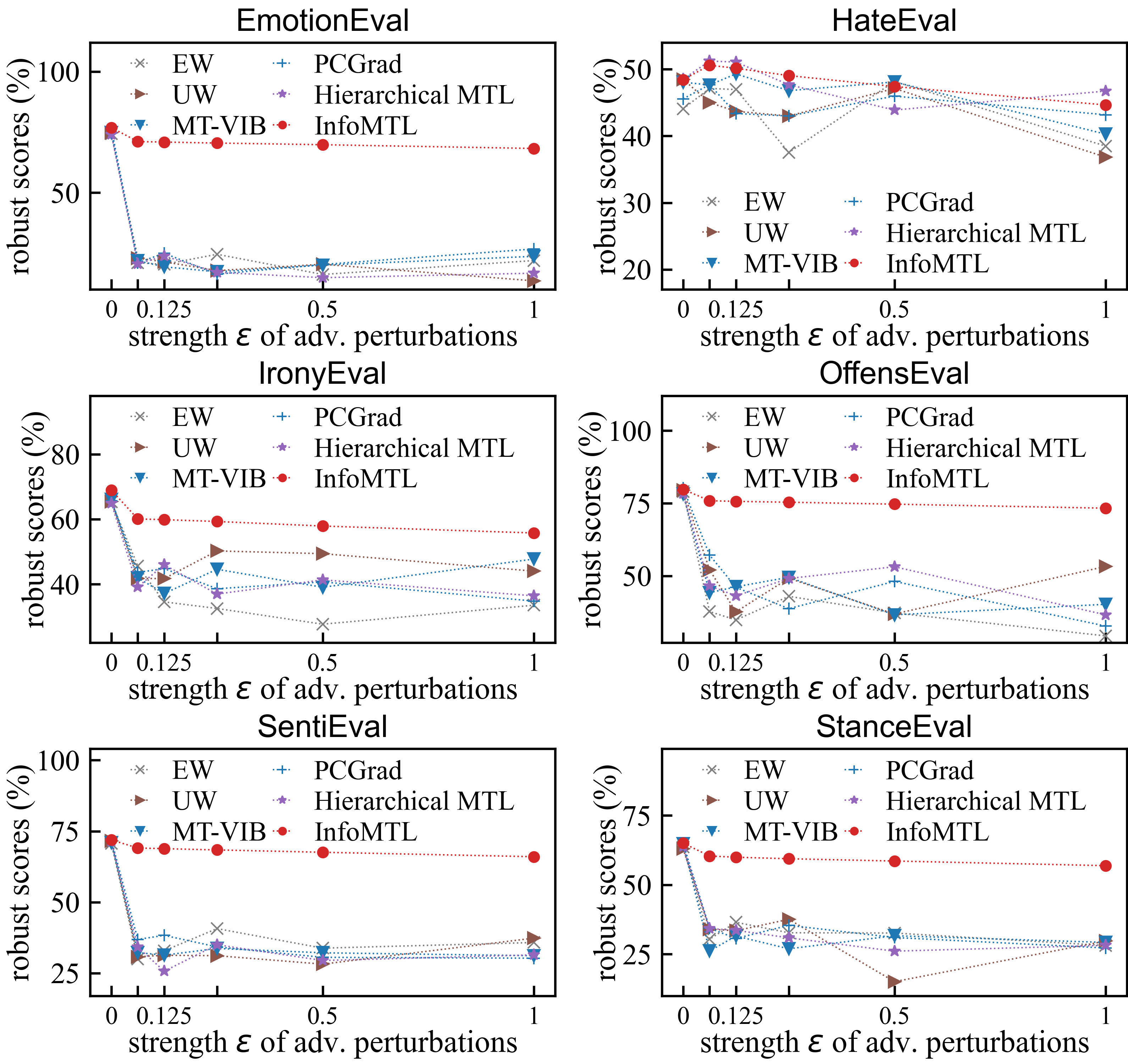}
       \caption{Robust scores (\%) against adversarial perturbation strengths. RoBERTa is the default backbone. }
    \label{fig:fgm_attack}
\end{figure}

\section{Conclusion}
We propose an information-theoretic multi-task learning framework (InfoMTL) to extract noise-invariant sufficient representations for all tasks, which can enhance language understanding of PLMs under the multi-task paradigm. It ensures sufficiency of shared representations for all tasks and mitigates the negative effect of redundant features. 
Firstly, the SIMax principle is proposed to learn more sufficient shared representations for all target tasks. In addition, the TIMin principle is designed to mitigate the negative effect of potential redundant features in the input for each task. 
Experiments on six benchmarks show that InfoMTL outperforms 12 comparative MTL methods under the same settings, especially in data-constrained and noisy scenarios. 

\appendix
\section*{Appendix Overview}
In this supplementary material, we provide: 
(i) the related work,
(ii) detailed description of experimental setups,
and (iii) supplementary results.

\section{Related Work} \label{sec:related}
Recent works on Multi-task Learning (MTL) mainly come from two aspects: multi-task optimization and network architecture design.
\subsection{Multi-task Optimization} 
The line of multi-task optimization works usually employ a hard parameter-sharing pattern \cite{DBLP:conf/icml/Caruana93}, where several light-weight task-specific heads are attached upon the heavy-weight task-agnostic backbone.

\paragraph{Task-balanced Methods} 
Task-balanced methods 
mainly focus on balancing learning process across multiple tasks for all tasks, such as loss-based and gradient-based methods.
Loss-based methods \cite{DBLP:conf/cvpr/KendallGC18,DBLP:conf/cvpr/ChennupatiSYR19,DBLP:conf/cvpr/LiuJD19,DBLP:conf/iclr/LiuLKXCYLZ21,DBLP:journals/tmlr/LinYZT22} focus on aligning the task loss magnitudes by rescaling loss scales. 
Gradient-based methods \cite{DBLP:conf/nips/SenerK18,DBLP:conf/icml/ChenBLR18,DBLP:conf/nips/YuK0LHF20} aims to find an aggregated gradient to balance different tasks.
Moreover, \citet{DBLP:conf/iclr/LiuLKXCYLZ21,DBLP:journals/tmlr/LinYZT22} provide general task balancing strategies that can simultaneously balance the loss and gradient of different tasks.
In real-world scenarios, the labeled data resource is limited and contains a certain amount of noise. The fact leads the above task-balanced MTL methods to perform suboptimally and struggle to achieve promising task prediction results.

\paragraph{Probabilistic Methods} 
Probabilistic methods
\cite{DBLP:conf/nips/YousefiSA19,DBLP:conf/iclr/KimCLH22,DBLP:journals/corr/abs-2007-00339,DBLP:conf/nips/ShenZWS21,DBLP:conf/ijcnn/FreitasBGM22} have been widely developed to explore shared priors for all tasks. The relationships among multiple tasks are investigated by designing priors over model parameters \cite{DBLP:conf/icml/YuTS05,DBLP:conf/nips/TitsiasL11,DBLP:conf/nips/ArchambeauGZ11,DBLP:journals/jmlr/BakkerH03} under the Bayesian framework, or sharing the covariance structure of parameters \cite{DBLP:conf/uai/Daume09}. 
In addition to the above methods that mainly focus on task relatedness or shared prior,
some works \cite{DBLP:journals/corr/abs-1711-07099,DBLP:journals/corr/abs-2007-00339,DBLP:conf/ijcnn/FreitasBGM22} introduce the information bottleneck (IB) principle \cite{tishby1999information,tishby2015deep} into the information encoding process of multi-task learning. These IB-based methods typically enhance the adaptability to noisy data by compressing task-irrelevant redundant information and learning compact intermediate representations. 
Specifically,
\citet{DBLP:journals/corr/abs-2007-00339} use the variational IB \cite{DBLP:conf/iclr/AlemiFD017} to learn probabilistic representations for multiple tasks.
\citet{DBLP:conf/ijcnn/FreitasBGM22} propose a hierarchical variational MTL method that restricts information individual tasks can access from a task-agnostic representation.

In multi-task scenarios, redundant information often differs across tasks, leading to situations where information beneficial for one task may become redundant for another. Directly applying the IB principle to compress redundancy for one task is prone to losing necessary information for other tasks. As a result, the learned shared representations would not only retain some redundant features but also face the task-specific insufficiency issue. To solve this, this paper proposes a new principled MTL framework InfoMTL to extract noise-invariant sufficient representations for all tasks. The proposed InfoMTL can ensure sufficiency for all target tasks and mitigate the negative effect of redundant features.

\subsection{Architectures for MTL} 
Orthogonal to the line of multi-task optimization, another line of MTL focuses on designing network architectures that mitigate task interference by optimizing the allocation of shared versus task-specific parameters \cite{DBLP:conf/cvpr/MisraSGH16,DBLP:conf/emnlp/HashimotoXTS17,DBLP:conf/aaai/RuderBAS19,DBLP:conf/cvpr/LiuJD19,DBLP:conf/acl/LiuHCG19}. Among them, some methods by soft parameter-sharing can share parameters among tasks to a large extent, but usually lead to high inference cost. 
The scope of our work is complementary to the architecture for MTL, as we mainly focus on learning better multi-task representations rather than designing better architectures.

\begin{table}[t]
\centering  
\small
\resizebox{0.99\linewidth}{!}{$
\begin{tabular}{l|c|rrr|r} 
\hline
\multicolumn{1}{c|}{\multirow{1}{*}{{Dataset}}} 
& \multicolumn{1}{c|}{\# {Label}}
& \multicolumn{1}{c}{\# {Train}} 
& \multicolumn{1}{c}{\# {Val}}  
& \multicolumn{1}{c|}{\# {Test}} 
& \multicolumn{1}{c}{\# {Total}} 
\\ 
\hline
EmotionEval  
& 4   &   3,257   &   374     &   1,421  &5,502 \\ 
HatEval  
& 2   &   9,000   &   1,000   &   2,970 & 12,970 \\ 
IronyEval   
& 2   &   2,862   &   955     &   784 & 4,601\\ 
OffensEval 
& 2   &   11,916  &   1,324   &   860 & 14,100 \\ 
SentiEval 
& 3   &   45,389  &   2,000   &   11,906 & 59,295\\ 
StanceEval  
& 3   &   2,620   &   294     &   1,249 & 4,163\\ 
\hline 
\end{tabular}
$}
\caption{Dataset statistics.}
\label{tab:datasets}
\end{table}

\begin{table}[t]
\centering
\small
\begin{tabular}{c|cc}
\hline 
\multicolumn{1}{c|}{\multirow{1}{*}{{Hyperparameter}}}  &
\multicolumn{1}{c}{BERT} & \multicolumn{1}{c}{RoBERTa} \\ \hline
Trade-off weight $\beta$  & 1 & 0.01  \\ 
Trade-off weight $\alpha$  & 0.01 & 0.1  \\  
Temperature $\tau$        & 0.1 & 1  \\
Number of epochs & 20 & 20  \\
Patience & 3 & 3\\ 
Batch size & 128 & 128   \\
Learning rate   & $5e^{-5}$ & $5e^{-5}$   \\ 
Weight decay & 0 & 0 
 \\ 
Dropout & 0 & 0.2  \\
Maximum token length & 256 & 256  \\
\hline 
\end{tabular}
\caption{Hyperparameters of InfoMTL.}
\label{tab:para}
\end{table}

\begin{table*}[t]
\centering
\small
\resizebox{0.93\linewidth}{!}{$
\begin{tabular}{l|cccccc|cc}
\hline
\multicolumn{1}{c|}{\multirow{2}{*}{{Methods}}} 
& \multicolumn{1}{c}{{EmotionEval}} 
& \multicolumn{1}{c}{{HatEval}} 
& \multicolumn{1}{c}{{IronyEval}} 
& \multicolumn{1}{c}{{OffensEval}} 
& \multicolumn{1}{c}{{SentiEval}} 
& \multicolumn{1}{c|}{{StanceEval}} 
& \multicolumn{1}{c}{\multirow{2}{*}{\textbf{Avg.}}} 
& \multicolumn{1}{c}{\multirow{2}{*}{$\mathbf{\Delta p} \uparrow$}} 
\\  
& M-F1 & M-F1 &  F1(i.) & M-F1 & M-Recall & M-F1 (a. \& f.) & \\
\hline 
\textbf{InfoMTL} & \textbf{76.90}\scriptsize{$\pm$0.62} & {48.44}\scriptsize{$\pm$2.15} & \textbf{68.94}\scriptsize{$\pm$1.86} & {79.78}\scriptsize{$\pm$0.86}  & \textbf{71.92}\scriptsize{$\pm$0.36} & {65.02}\scriptsize{$\pm$1.81} & \textbf{68.50}\scriptsize{$\pm$0.58} & \textbf{+3.97} \\
\ \ w/o SIMax & 76.02\scriptsize{$\pm$1.10} & \textbf{49.30}\scriptsize{$\pm$3.75}  & 64.63\scriptsize{$\pm$3.14} & 79.44\scriptsize{$\pm$1.93}  & 71.67\scriptsize{$\pm$1.25} & 64.47\scriptsize{$\pm$1.65} & 67.59\scriptsize{$\pm$1.06} & +2.72 \\ 
\ \ w/o TIMin & 76.37\scriptsize{$\pm$0.37} & 46.82\scriptsize{$\pm$4.98}  & 64.12\scriptsize{$\pm$7.01} & \textbf{80.25}\scriptsize{$\pm$0.89} & 71.39\scriptsize{$\pm$0.49} & \textbf{66.34}\scriptsize{$\pm$0.74} & 67.55\scriptsize{$\pm$0.31} & +2.34 \\ 
\ \ w/o SIMax \& TIMin
& {74.37}\scriptsize{$\pm$0.56} & {44.08}\scriptsize{$\pm$5.26} & {65.32}\scriptsize{$\pm$1.84} & {79.04}\scriptsize{$\pm$1.43} & {70.64}\scriptsize{$\pm$1.71} & {63.59}\scriptsize{$\pm$2.43} & {66.17}\scriptsize{$\pm$0.43} & 0.00 \\
\hline
\end{tabular}
$}
\caption{Fin-grained ablation results (\%) of  InfoMTL with RoBERTa backbone. }
\label{tab:fine_results_abla}
\end{table*}

\section{Experimental Setups}
\paragraph{Datasets and Downstream Tasks}
This study primarily focuses on MTL in the field of natural language understanding, and proposes a new MTL approach to better handle real-world scenarios with data noise and limited labeled data. To effectively validate the proposed method, we selected six text classification benchmarks \cite{DBLP:conf/emnlp/BarbieriCAN20} from social media, which naturally contain some noise, to evaluate multi-task performance.
The statistics are listed in Table~\ref{tab:datasets}.
\textit{EmotionEval} \cite{DBLP:conf/semeval/MohammadBSK18} involves detecting the emotion evoked by a tweet and is based on the Affects in Tweets conducted during SemEval-2018.
Following \citet{DBLP:conf/emnlp/BarbieriCAN20,DBLP:conf/aaai/0001WLZH24}, the four most common emotions (i.e., anger, joy, sadness, and optimism) are selected as labels.
\textit{HatEval} \cite{DBLP:conf/semeval/BasileBFNPPRS19} stems from SemEval-2019 HatEval challenge and is used to predict whether a tweet is hateful towards immigrants or women.
\textit{IronyEval} \cite{DBLP:conf/semeval/HeeLH18} is from SemEval-2018 Irony Detection and consists of identifying whether a tweet includes ironic intents or not.
\textit{OffensEval} \cite{DBLP:conf/semeval/ZampieriMNRFK19} is from SemEval-2019 OffensEval and involves predicting whether a tweet contains any form of offensive language. 
\textit{SentiEval} \cite{DBLP:conf/semeval/RosenthalFN17} comes from SemEval-2017 and is designed for the task of determining whether a tweet expresses a positive, negative, or neutral sentiment.
\textit{StanceEval} \cite{DBLP:conf/semeval/MohammadKSZC16} involves determining if the author of a piece of text has a favorable, neutral, or negative position towards a proposition or target. 

\paragraph{Description of Comparison Methods}
we compare with the following 12 representative MTL methods. \textbf{Equal Weighting} (EW) is a typical baseline that applies equal weights for each task.
MT-DNN \cite{DBLP:conf/acl/LiuHCG19} is a version of EW baseline with the BERT backbone.
\textbf{Task Weighting} (TW) assigns loss weights to each task based on the ratio of task examples. 
\textbf{Scale-invariant Loss} (SI) is invariant to rescaling each loss with a logarithmic operation.
\textbf{Uncertainty Weighting} (UW) \cite{DBLP:conf/cvpr/KendallGC18} uses the homoscedastic uncertainty quantification to adjust task weights.
\textbf{Geometric Loss Strategy} (GLS) \cite{DBLP:conf/cvpr/ChennupatiSYR19} uses the geometric mean of task losses to the weighted average of task losses.
\textbf{Dynamic Weight Average} (DWA) \cite{DBLP:conf/cvpr/LiuJD19}  sets the loss weight of each task to be the ratio of two adjacent losses.
\textbf{Projecting Conflicting Gradient} (PCGrad) \cite{DBLP:conf/nips/YuK0LHF20}  removes conflicting components of each gradient w.r.t the other gradients.
\textbf{IMTL-L} \cite{DBLP:conf/iclr/LiuLKXCYLZ21} dynamically reweighs the losses such that they all have the same magnitude.
\textbf{Random Loss Weighting} (RLW) \cite{DBLP:journals/tmlr/LinYZT22} with normal distribution, scales the losses according to randomly sampled task weights.
\textbf{MT-VIB} \cite{DBLP:journals/corr/abs-2007-00339}
is a variational MTL method based on information bottleneck. 
\textbf{VMTL}
\cite{DBLP:conf/nips/ShenZWS21} is a variational MTL framework that uses Gumbel-Softmax priors for both representations and weights.
\textbf{Hierarchical MTL}
\cite{DBLP:conf/ijcnn/FreitasBGM22}
is a hierarchical variational MTL method with compressed task-specific representations based on information bottleneck.
We also compare with GPT-3.5, an enhanced generative pre-trained transformer model based on text-davinci-003\footnote{We present the zero-shot results of the GPT-3.5-turbo snapshot from June 13th 2023.}, optimized for chatting.

\begin{figure}[t]
    \centering
    \includegraphics[width=.99\linewidth]{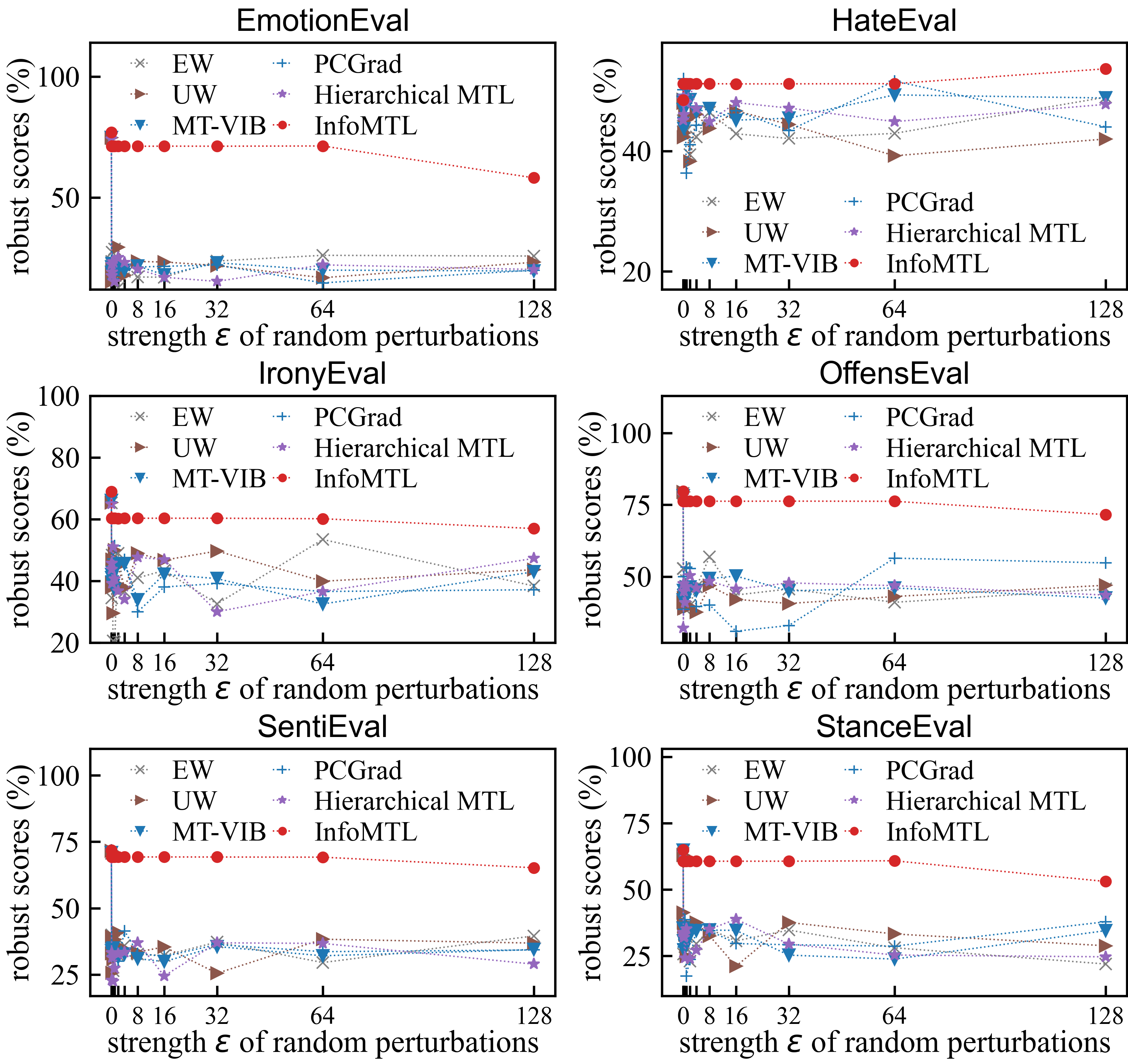}
    \caption{Robust scores (\%) against random perturbation strengths. RoBERTa is the default backbone.     }
    \label{fig:random_attack}
\end{figure}

\paragraph{Details of Evaluation Metrics}
Following \citet{DBLP:conf/emnlp/BarbieriCAN20,DBLP:conf/aaai/0001WLZH24}, the global metric based on the average of all dataset-specific metrics, computed as ${\mathbf{Avg.}}=\frac{1}{|T|}\sum_{t=1}^{|T|} \frac{1}{N_t}\sum_{n-1}^{N_t} M_{t,n},$
where $M_{t,n}$ denotes the performance for the $n$-th metric in task $t$. $N_t$ denotes the number of metrics in task $t$. $|T|$ refers to the number of tasks.
Following \citet{DBLP:conf/cvpr/ManinisRK19,DBLP:journals/tmlr/LinYZT22}, the average relative improvement of each method over the EW baseline as the multi-task performance measure, denoted as 
$\mathbf{\Delta p}=\frac{1}{|T|}\sum_{t=1}^{|T|} \frac{1}{N_t}\sum_{n-1}^{N_t} \frac{(-1)^{p_{t,n}}(M_{t,n}-M^{EW}_{t,n})}{M^{EW}_{t,n}},$
where $M^{EW}_{t,n}$ is the $n$-th metric score for EW on task $t$.
$p_{t,n} = 0 $ if a higher value is better for the $n$-th metric in task $t$ and 1 otherwise.

\paragraph{Implementation Details}
Table~\ref{tab:para} shows the best parameters of our InfoMTL with RoBERTa and BERT backbones.

\section{Supplementary Results}
Table~\ref{tab:fine_results_abla} presents the fine-grained ablation results.
Figure~\ref{fig:random_attack} illustrates the results against random perturbation strengths.

\section*{Acknowledgements} 
This work was supported by the National Natural Science Foundation of China (No. U24A20335), and the Postdoctoral Fellowship Program of China Postdoctoral Science Foundation (No. GZC20232969).
The authors thank the anonymous reviewers and the meta-reviewer for their helpful comments on the paper.

\bibliography{aaai25}

\end{document}